\documentclass[final,5p,times,twocolumn]{elsarticle}

\usepackage{graphicx}
\usepackage{amssymb}
\usepackage{pdflscape}
\usepackage[english]{babel}
\usepackage[dvipsnames,svgnames,x11names]{xcolor}
\usepackage{booktabs,tabularx}
\usepackage{multirow}
\usepackage{url}
\usepackage{amsmath}
\usepackage{commath}
\usepackage[english]{babel}
\usepackage[ruled]{algorithm2e}
\SetEndCharOfAlgoLine{}
\usepackage[leftmargin=6em,rightmargin=12em,indentfirst=false]{quoting}
\usepackage{siunitx}

\usepackage[final]{changes}
\usepackage{float}

\usepackage{lineno}
\usepackage{tikz}
\usepackage[export]{adjustbox}

\usepackage{graphicx}
\usepackage{caption}
\usepackage{subcaption}
\usepackage[utf8]{inputenc}
\usepackage[export]{adjustbox}
\usepackage{wrapfig}
\usepackage{dcolumn}
\usepackage{subfig}
\usepackage{multirow}
\usepackage{multicol}

\makeatletter
\newcolumntype{T}[3]{>{\textfont0=\the@{#1}{#2}{#3}}c<{\DC@end}}
\makeatother

\usepackage{pgfplots}
\pgfplotsset{width=10cm,compat=1.9}

\usepackage{array}

\newcolumntype{L}[1]{>{\raggedright\let\newline\\\arraybackslash\hspace{0pt}}m{#1}}
\newcolumntype{C}[1]{>{\centering\let\newline\\\arraybackslash\hspace{0pt}}m{#1}}
\newcolumntype{R}[1]{>{\raggedleft\let\newline\\\arraybackslash\hspace{0pt}}m{#1}}

\usepackage{subfiles}

\usepackage[utf8]{inputenc}
\usepackage[T1]{fontenc}
\usepackage{lineno}
\usepackage{amsmath}
\usepackage{gensymb}
\usepackage{threeparttable}
\usepackage{graphicx}
\usepackage{multirow}
\usepackage{makecell}
\usepackage[nameinlink,capitalise]{cleveref}

\usepackage{todonotes}

\journal{Building and Environment}
\begin{document}



\title{Semantic segmentation of longitudinal thermal images for identification of hot and cool spots in urban areas}
\author[inst1]{Vasantha Ramani}

\affiliation[inst1]{organization={Berkeley Education Alliance for Research in Singapore},
            addressline={CREATE Tower 1 Create Way}, 
           postcode={138602}, 
            country={Singapore}}

\affiliation[inst2]{organization ={Robert Bosch Centre for Cyber-physical Systems, Indian Institute of Science}, 
    addressline = {Bengaluru, Karnataka}, 
    postcode={560012}, 
    country={India}}
    
\affiliation[inst3]{organization={Department of Electrical Engineering and Computer Sciences, University of California},
            addressline={Berkeley}, 
           postcode={94720-1770}, 
            country={CA}}
            
\affiliation[inst4]{organization={Department of the Built Environment, College of Design               and Engineering, National University of Singapore},
           addressline={4 Architecture Drive}, 
           postcode={117566}, 
            country={Singapore}}

\author[inst1,inst2]{Pandarasamy Arjunan}
\author[inst3]{Kameshwar Poolla}
\author[inst4]{Clayton Miller\corref{cor1}}
\ead{clayton@nus.edu.sg}
\cortext[cor1]{Corresponding author}


\begin{abstract}
This work presents the analysis of semantically segmented, longitudinally, and spatially rich thermal images collected at the neighborhood scale to identify hot and cool spots in urban areas. An infrared observatory was operated over a few months to collect thermal images of different types of buildings on the educational campus of the National University of Singapore. A subset of the thermal image dataset was used to train state-of-the-art deep learning models to segment various urban features such as buildings, vegetation, sky, and roads. It was observed that the U-Net segmentation model with `resnet34' CNN backbone has the highest \textit{mIoU} score of 0.99 on the test dataset, compared to other models such as DeepLabV3, DeeplabV3+, FPN, and PSPnet. The masks generated using the segmentation models were then used to extract the temperature from thermal images and correct for differences in the emissivity of various urban features. Further, various statistical measures of the temperature extracted using the predicted segmentation masks are shown to closely match the temperature extracted using the ground truth masks. Finally, the masks were used to identify hot and cool spots in the urban feature at various instances of time. This forms one of the very few studies demonstrating the automated analysis of thermal images, which can be of potential use to urban planners for devising mitigation strategies for reducing the urban heat island (UHI) effect, improving building energy efficiency, and maximizing outdoor thermal comfort.  
\end{abstract}




\maketitle

\section{Introduction}
Over 50\% of the world population resides in urban areas, which is expected to increase in the future (\cite{ritchie2018urbanization}). As a result, several urban cities face the challenge of increased demand for various resources such as energy, housing, transportation, and health care (\cite{zhang2016trends}). In addition to the rising population, climate change can exacerbate the existing problems in the urban environment and may affect the quality of life of urban dwellers. Hence, understanding the dynamics of the urban environment is crucial to designing it in such a way as to meet the needs of the growing population and also implement suitable mitigation measures to tackle the impacts of climate change (\cite{manoli2019magnitude}). 

Imaging is a commonly used non-contact technique for conducting urban planning studies on land use patterns, urban heat island (UHI) effect, built environment, pedestrian detection, surveillance, and many more (\cite{tamiminia2020google, coutts2016thermal, adao2017hyperspectral}). It can be carried out at different scales, such as the microscale using hand-held cameras, the local scale using drones and observatories, and the macroscale using aerial vehicles and satellites (\cite{martin2022infrared}). The type of application depends not just on the scale at which it is carried out but also on the light spectrum range that is being captured, such as visible, infrared, hyperspectral, and multi-spectral.

Of the different spectral imaging, infrared imaging can capture the objects' long-wave infrared wavelengths, which indicates their surface temperature. It has been used for studying various aspects of the urban environment, such as identification of damage in the built environment, urban heat islands, extracting the thermal properties of the building components, and characterizing HVAC usage patterns (\cite{martin2022infrared, ramani2023longitudinal}). The processing of thermal images is one of the essential steps in extracting temperature and object information. Segmentation of images based on object/feature type is an image processing step that allows pixel-wise labeling of the images.  

Segmentation of RGB/visible images has been demonstrated extensively; however, very limited studies have been carried out on the segmentation of thermal images (\cite{kutuk2022semantic}). This is mainly due to the lack of spatially and longitudinally rich urban-scale thermal images and difficulties associated with the segmentation of thermal images. This paper addresses this research gap through an urban-scale study involving semantic segmentation of longitudinal thermal images captured at the neighborhood scale. Specifically, we employed the U-net deep learning architecture for the semantic segmentation of urban features in the IRIS dataset. This model achieved the highest mean intersection over union (\textit{mIoU}) score of 0.99 on a test set compared to four other contemporary deep learning architectures. The segmented images are then used to identify hot and cool spots in the urban environment. Studying the diurnal changes in the hot and cool spots' locations at the neighborhood scale can assist urban planners in devising suitable mitigation measures to reduce the UHI effect and improve building energy efficiency in urban areas.

The subsequent section presents a background on the semantic segmentation of thermal images. Section~\ref{sec:methodology} describes the methodology adopted for collecting thermal images, their segmentation, and analysis. In Section ~\ref{sec:results}, the segmentation and thermal analysis results are discussed in detail. Finally, in Sections~\ref{sec:discussion} and ~\ref{sec:conclusions}, the discussion and conclusions of the study are presented. 

\section{Related studies}
Segmentation of images is an essential tool that involves the assignment of a label to each pixel in an image. Segmentation of images is used for various applications such as medical imaging and diagnosis, autonomous driving, pedestrian detection, and others (\cite{khan2021deep, feng2020deep, xiao2021deep, garcia2017review, atif2019review}). Some of the commonly used segmentation methods are threshold-based segmentation (\cite{bhargavi2014survey}), segmentation based on edge detection (\cite{dhankhar2013review, savant2014review}) and clustering (\cite{sharma2016review}), region-based segmentation and more recently segmentation using deep learning methods (\cite{minaee2021image, garcia2017review}). With the advancement in computational capacity and robust segmentation models, deep learning has been the most sought-after method for the semantic segmentation of digital images.

Several types of deep learning methods for segmentation have been developed and tested on RGB datasets. More recently, segmentation using an encoder for down-sampling and a decoder for up-sampling has achieved promising results. Some of the models include DeepLabV3 (\cite{chen2017rethinking}), DeepLabV3+ (\cite{chen2018encoder}), U-net (\cite{ronneberger2015u}), FPN (\cite{lin2017feature}), PSPNet (\cite{zhao2017pyramid}) and others. The performance of these models has been tested on various digital image datasets such as Cityscapes (\cite{cordts2016cityscapes}), ImageNet (\cite{deng2009imagenet}), and others. Though these models have been extensively used for digital images, very limited studies have been conducted on the segmentation of infrared images.

The advantage of using infrared images is that capturing information even in low illumination is possible, which may not be possible using RGB images. However, the thermal images have low resolution and ambiguous boundaries between various objects. In addition, unlike the RGB dataset, there is a lack of an annotated infrared image dataset (\cite{kutuk2022semantic}). Some of the thermal datasets include the `Segmenting objects in Day and Night'(SODA) dataset (\cite{li2020segmenting}), OSU thermal pedestrian dataset (\cite{miezianko2008people}), SCUT-Seg dataset (\cite{XIONG2021103628}), and \texttt{NPU\_CS\_UAV\_IR\_DATA} (\cite{liu2018real}). These datasets are from varied urban scenes and are mainly used for pedestrian detection and autonomous driving applications.

Due to the lack of a thermal image dataset, very few segmentation models have been developed exclusively for thermal images. Nevertheless, segmentation of both multi-spectral (RGB and infrared) (\cite{ha2017mfnet, shivakumar2020pst900, sun2019rtfnet, vertens2020heatnet}) and infrared only (\cite{li2020segmenting, wang2019thermal, panetta2021ftnet}) images have been demonstrated in recent past. The neural network models developed for the segmentation of RGB images have been modified and adapted for the segmentation of thermal images. The multi-spectral segmentation models take RGB and thermal images as input. It consists of two encoders to take RGB and thermal images as input and one decoder for upscaling. However, for multi-spectral segmentation, it is required to have spectrally aligned images and special hardware systems that can handle both infrared and visible images. This requirement can be avoided in the case of infrared-only images.

The IRIS dataset (\cite{lin2023district}) is one of the very few datasets that consists of longitudinally and spatially rich thermal images of the built environment at the neighborhood scale. Semantic segmentation of such a dataset can be of potential use for urban-scale studies on the urban heat island effect, vegetation, traffic, and building space. In this paper, the performance of various state-of-the-art deep learning encoder-decoder type segmentation models is explored to study the temporal and spatial variations in the temperature of the urban feature.

\section{Methodology}
\label{sec:methodology}
\begin{figure*}
  \centering
  \includegraphics[width=.9\linewidth]{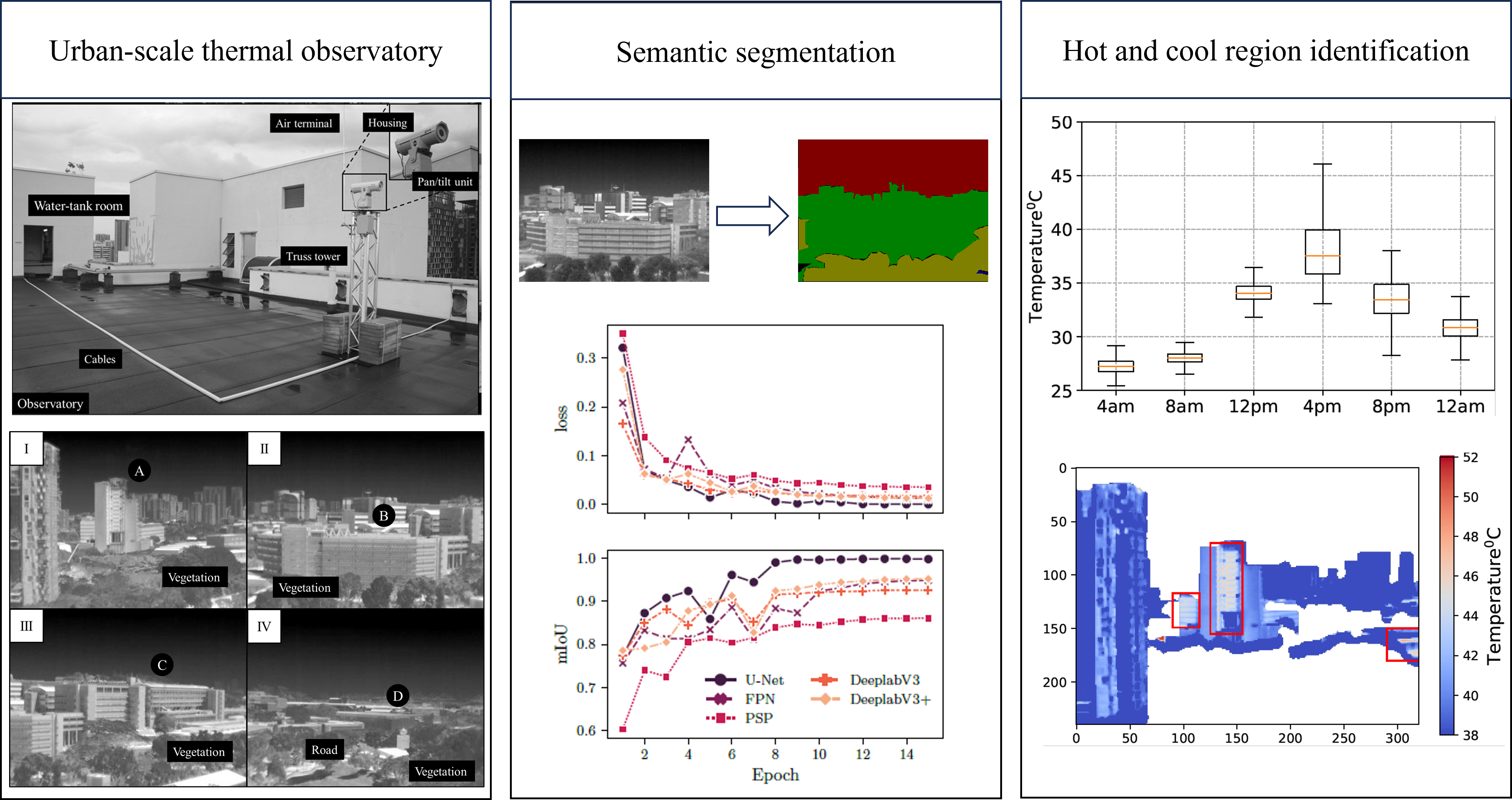}
  \caption{Schematic showing the methodology adopted for analysis of thermal images to identify various urban features and to study the temperature distribution within various urban features.}.
  \label{fig:methodology}
\end{figure*}

Figure \ref{fig:methodology} shows the overview of the methodology adopted for the segmentation and analysis of the thermal images using deep learning segmentation models to identify various urban features. This section discusses each of the elements demonstrated in the figure in detail.

\subsection{IR observatory and thermal image dataset}
\begin{figure}
  \centering
  \includegraphics[width=\linewidth]{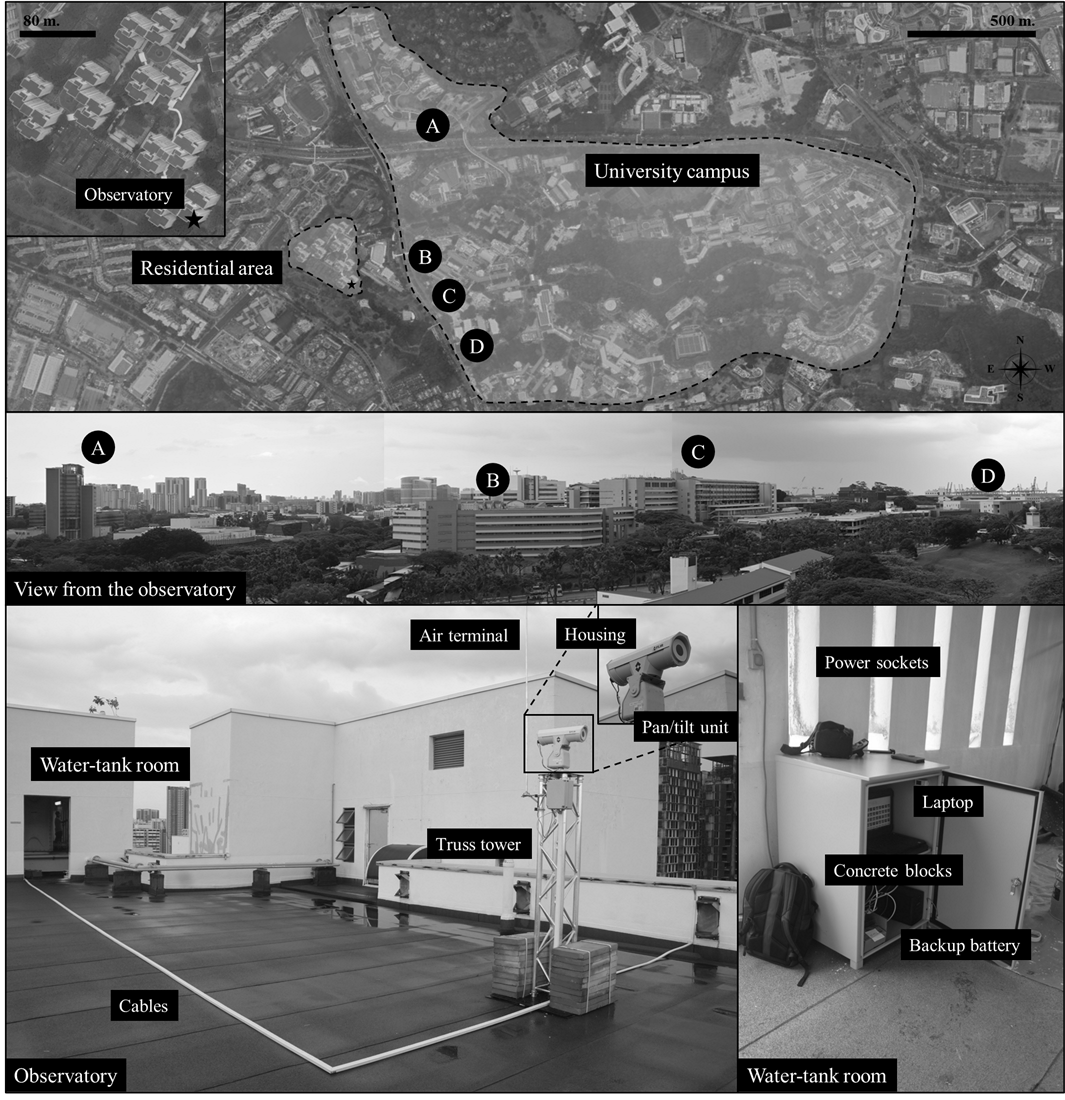}
  \caption{Google map showing the location of the IR observatory for deployment 1 (top), digital image of the buildings, which were imaged using the thermal camera (center), and the infrared observatory (bottom) \cite{martin2022iris}}.
  \label{fig:observatory and location kv}
\end{figure}

\begin{figure}
  \centering
  \includegraphics[width=\linewidth]{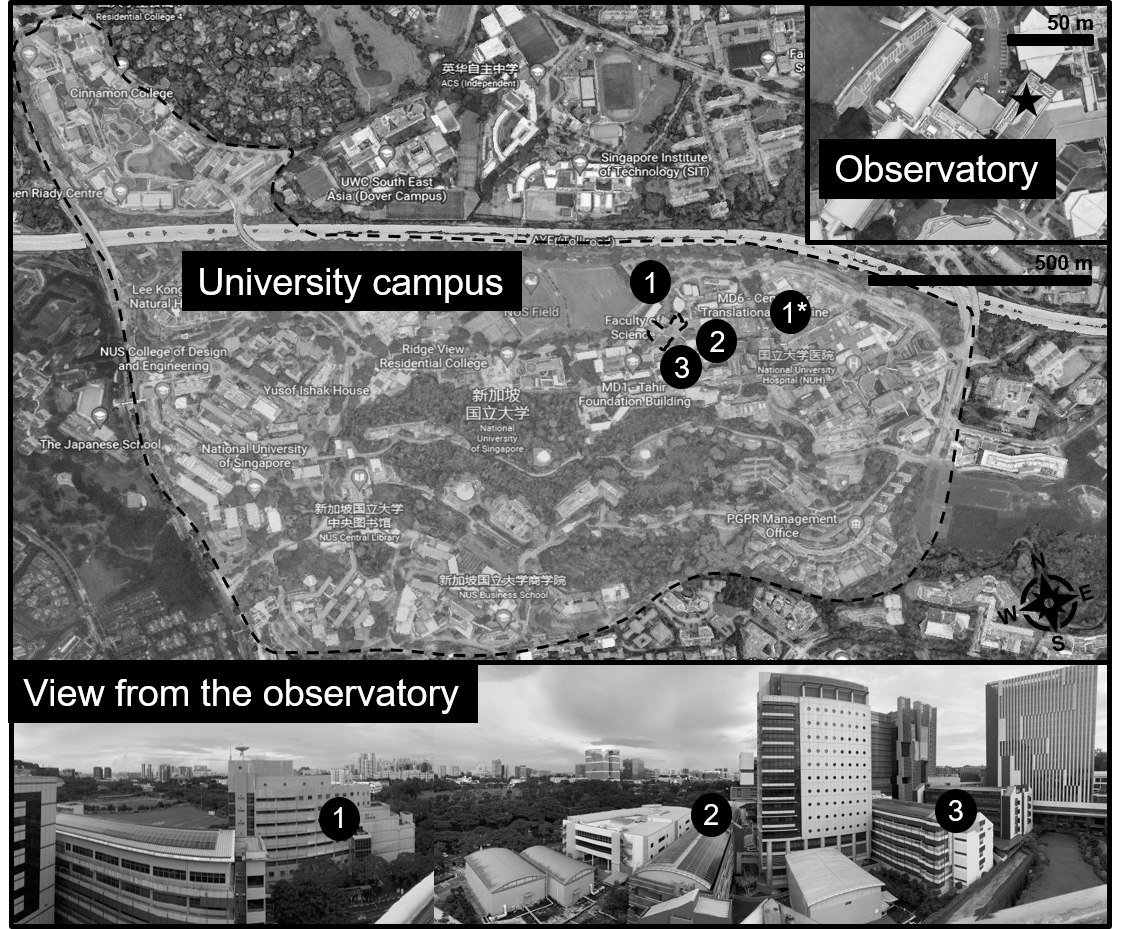}
  \caption{Google map showing the location of the IR observatory for deployment 2 (top), and the buildings imaged using the thermal camera (bottom) \cite{lin2023district}}.
  \label{fig:location s16}  
\end{figure}

A neighborhood-scale observatory, as shown in Figure \ref{fig:observatory and location kv}, was installed to capture thermal images of buildings on the educational campus of the National University of Singapore (NUS). Two installations were made, and the corresponding locations and the digital images of the buildings observed during the period of operation are shown in Figure \ref{fig:observatory and location kv} (location 1) and Figure \ref{fig:location s16} (location 2) respectively. The observatory was operated at Location 1 from November 2021 to March 2022 and at Location 2 from August to December 2022.  

These locations were selected to generate a longitudinally and spatially rich thermal image dataset with diverse building types. For instance, in Figure \ref{fig:observatory and location kv}, Building A is partially glazed, while Building B and C are reinforced concrete buildings, and Building D is a net zero building. In addition to the different building types, urban features such as traffic and vegetation were also captured using the thermal camera. As shown in Figure \ref{fig:observatory and location kv}, the thermal camera was housed inside a protective casing mounted on a pan-tilt unit. The pan-tilt unit was allowed to rotate along the horizontal axis to capture images of various buildings. Readers can refer to the paper by \cite{lin2023district} for a detailed description of the installation and the different views captured using the thermal camera. In the subsequent sections, we refer to Views A, B, C, and D of Location 1 as Views 1, 2, 3, and 4, respectively, and Views 1, 2, and 3 of Location 2 as Views 5, 6, and 7, respectively. 

The images were captured using a FLIR A300 thermal camera, whose specifications are listed in Table \ref{tab:camera specs}. The captured images were transferred to the cloud and stored for further analysis. The thermal images were segregated into different views using a convolution neural network (CNN) model described in \cite{ramani2023longitudinal}. After collecting thermal images, part of the dataset was used for training the segmentation model. The methodology for segmentation and model training is described in the subsequent section.

\begin{table}
  \centering
  \caption{FLIR A300 Thermal camera specifications}
  \label{tab:camera specs}
  \begin{tabular}{cc}
    \toprule
    Resolution & 16 bit, 320x240 pixels\\
    Thermal sensitivity & 50mK $@$ 30$^o$C\\
    Sensor & Uncooled Microbolometer FPA\\ 
    Spectral range & 7.5 to 13 $\mu$ m\\ 
    Field of view (FOV) & 25$^o$ (H) and 18.8$^o$ (V)\\
    Accuracy & $\pm$2$^o$ or 2\% of reading\\
    Power supply & 110/220 V AC\\
    Weight & 0.7 kg\\
    Size & 170mm x 70mm x 70mm\\
  \bottomrule
\end{tabular}
\end{table}

\begin{table}
  \centering
  \caption{FLIR A300 Thermal camera default Plank's constants}
  \label{tab:camera constants}
  \begin{tabular}{ccl}
    \toprule
    \emph{R\textsubscript{1}} & 14911.1846\\
    \emph{R\textsubscript{2}} & 0.0108\\
    \emph{f} & 1.0\\ 
    \emph{O} & -6303.0\\ 
    \emph{B} & 1396.6\\
  \bottomrule
\end{tabular}
\end{table}

\begin{figure}
  \centering
  \includegraphics[width=.8\linewidth]{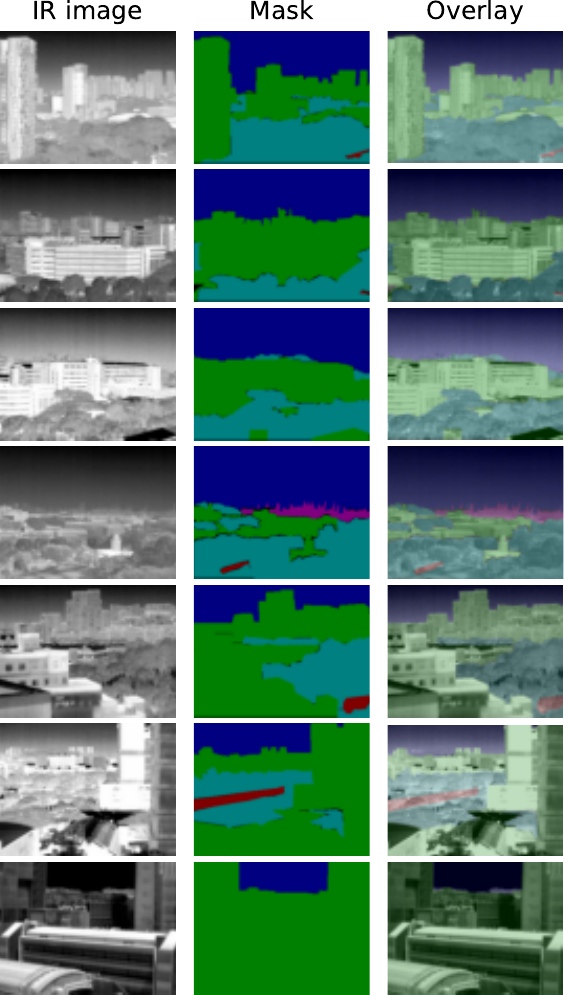}
  \caption{The thermal image, the ground truth mask generated using the LabelMe annotation tool, and the overlay image of the different views captured using the thermal camera. The urban features: buildings, vegetation, road, sky, and offshore structures are displayed in green, cyan, blue, red, and pink color respectively.}
  \label{fig:mask_and_overlay}
\end{figure}

\subsection{Semantic Segmentation of IR images}
The first step in the semantic segmentation of images using deep learning involves the preparation of the dataset. A sample image from each view was segmented using the Labelme (\cite{torralba2010labelme}) annotation tool. This was achieved by comparing the thermal and digital images and marking the regions corresponding to different urban features. Each pixel value in the masked image corresponds to one of the urban features, such as buildings, vegetation, road, sky, or offshore structures. Any pixel not corresponding to one of the features is labeled as a background pixel. Figure \ref{fig:mask_and_overlay} shows the thermal image, the corresponding masked image, and the overlay image for some locations. A total of 8,953 images were stratified and sampled from the IRIS dataset based on location and view and across several days. These IR images and their corresponding masks were used for training and testing the neural network segmentation model.

Following the generation of ground truth masks, the images were transformed such that the pixel values were in the range of [0,1] and subsequently normalized with a mean = [0.485, 0.485, 0.485] and standard deviation = [0.229, 0.229, 0.229]. In this study, we selected five state-of-the-art deep learning-based image segmentation models: U-Net (\cite{ronneberger2015u}), Feature Pyramid Network or FPN (\cite{lin2017feature}), Pyramid scene parsing network or PSPNet (\cite{zhao2017pyramid}), DeepLabv3(~\cite{chen2017rethinking}), and DeepLabv3+(~\cite{chen2017rethinking}). The following models were selected for their significant advancements in image segmentation and wide adoption in various computer vision tasks:\footnote{Our objective is to leverage existing deep learning-based segmentation models for analyzing urban heat/cool spots. Improving or analyzing these models' characteristics falls outside this study's scope.}

\begin{figure}
  \centering
  \includegraphics[width=.9\linewidth]{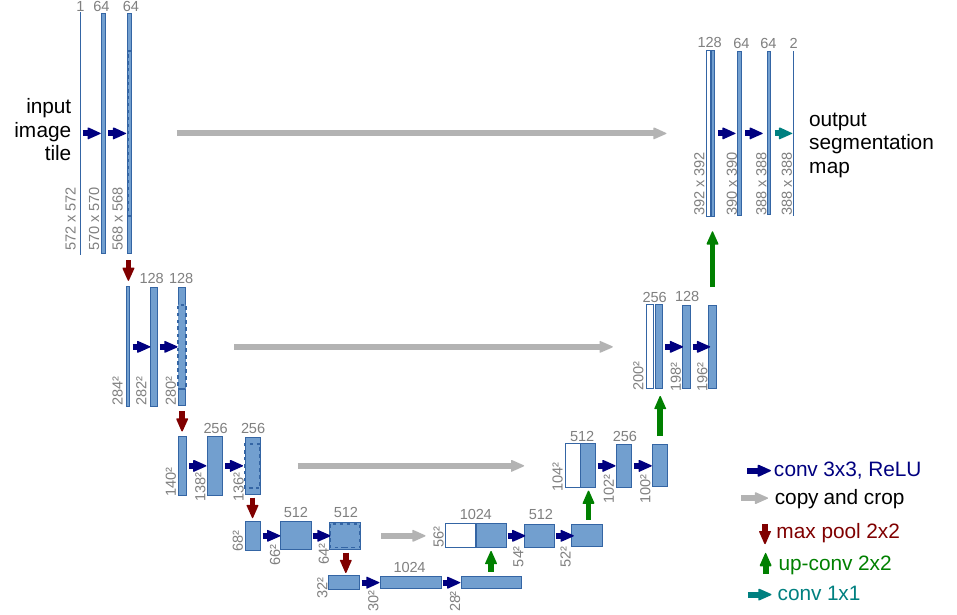}
  \caption{U-Net architecture \cite{ronneberger2015u}.}
  \label{fig:unet}
\end{figure}

\subsubsection{U-Net}
U-Net (\cite{ronneberger2015u}) is a convolutional neural network (CNN) architecture primarily employed for image segmentation tasks in various image analysis applications. The name 'U-Net' is coined from the network's characteristic U-shaped architecture, as shown in Figure~\ref{fig:unet}. The network has two paths: a contracting path on the left and an expansive path on the right. The contracting path has two 3x3 convolutions (unpadded) and a rectified linear unit (ReLU) after each one, and a 2x2 max pooling with stride 2 for downsampling. The number of feature channels doubles at each downsampling step. The expansive path has an upsampling of the feature map and a 2x2 convolution ("up-convolution") that halves the number of feature channels, a concatenation with the cropped feature map from the contracting path, and two 3x3 convolutions with a ReLU after each one. The cropping is needed because the convolutions reduce the border pixels. The last layer has a 1x1 convolution that maps each 64-component feature vector to the number of classes. The network has 23 convolutional layers in total (\cite{ronneberger2015u}).

The U-Net architecture is renowned for its unique ability to achieve precise and pixel-wise segmentation of images, ensuring that the input and output share the same dimensions. Furthermore, U-Net's extensive use of skip connections allows it to capture both high-level and low-level features, preserving valuable spatial information that might otherwise be lost during down-sampling operations in the encoder. These distinctive features make it exceptionally valuable for tasks that demand accurate delineation of object boundaries and shapes within an image. 

\subsubsection{Feature Pyramid Network (FPN)} FPN (\cite{lin2017feature}) is a versatile feature extractor that generates proportionally sized feature maps at multiple levels from a single-scale input image, independently of the underlying convolutional architectures. This makes FPN a generic solution for constructing feature pyramids within deep convolutional networks, particularly beneficial for tasks like object detection.
The pyramid construction involves a bottom-up pathway and a top-down pathway. The bottom-up pathway is driven by the feedforward computation of the backbone ConvNet, producing feature hierarchies at various scales. The top-down pathway upsamples spatially coarser feature maps from higher pyramid levels, enhancing them with semantically stronger features. Lateral connections merge feature maps from the bottom-up and top-down pathways, combining localized accuracy from the former with higher-level semantics from the latter. This integration enhances the network's ability to capture information at different scales, contributing to improved performance in tasks such as object detection (\cite{lin2017feature}).


\subsubsection{Pyramid Scene Parsing Network (PSPNet)} 
PSPNet (\cite{zhao2017pyramid}) is a semantic segmentation model employing a pyramid parsing module to leverage global context information through region-based context aggregation. By combining local and global clues, the PSPNet enhances the reliability of its final predictions. It use a pre-trained CNN with a dilated network strategy on the input image, resulting in a final feature map of the same size as the input. The pyramid pooling module is then applied to this map, encompassing whole, half, and smaller image portions. The fused information serves as a global prior, concatenated with the original feature map, and processed through a convolution layer to produce the final prediction map.


\subsubsection{DeepLabv3} 
DeepLabv3 (\cite{chen2017rethinking}) is a semantic segmentation architecture that builds upon DeepLabv2 (\cite{chen2017deeplab}), incorporating several key modifications. Specialized modules are introduced to address the challenge of segmenting objects across multiple scales. These modules utilize atrous convolution either sequentially or in parallel, employing multiple atrous rates to capture diverse multi-scale contexts. Additionally, the Atrous Spatial Pyramid Pooling (ASPP) module from DeepLabv2 is enhanced by integrating image-level features to augment global context understanding, thereby improving overall performance.
Notable changes to the ASPP module involve applying global average pooling to the final feature map of the model. The resulting image-level features undergo processing through a 1 × 1 convolution with 256 filters (along with batch normalization). Subsequently, the features are bilinearly upsampled to attain the desired spatial dimension. The refined ASPP module now comprises one 1×1 convolution and three 3 × 3 convolutions with atrous rates set to (6, 12, 18) when the output stride is 16. Each convolution has 256 filters and incorporates batch normalization. Additionally, the image-level features contribute to the improved ASPP, enriching the network's ability to capture contextual information across different scales.

\subsubsection{DeepLabv3+}
DeepLabv3+ (\cite{chen2017rethinking}) is an extension of DeepLabv3 that includes an additional decoder module to refine the segmentation results. Moreover, the Xception model is adapted for the segmentation task, and depthwise separable convolution is applied to both the Atrous Spatial Pyramid Pooling and decoder modules, resulting in a faster and more robust encoder-decoder network.

The segmentation models were initialized with pre-trained weights on the ImageNet dataset (\cite{deng2009imagenet}). A hold-out procedure with a test-to-train ratio of 0.2 and a validation-to-train ratio of 0.25 was used to train the segmentation model and subsequently test its performance. The segmentation models were implemented on the PyTorch platform. The model weights were estimated such that the trained model yielded a minimum error between the ground truth and the predicted masks. This work used cross-entropy loss functions with equal weights for all classes to estimate the model performance in the training phase. The losses were estimated for model weights optimized using Adam optimizer. The models were evaluated using mean intersection over union (\textit{mIoU}) (\cite{he2016deep}) metric on the validation, test, and train dataset, defined as follows:

 \begin{equation}
 \label{eq: IoU}
  mIoU = \frac{1}{k}.\sum_{n=0}^{k}\frac{Area_{intersection}}{Area_{union}}
 \end{equation}
  where $Area_{intersection}$ is the area of intersection and $Area_{union}$ is the area of union between the predicted and the ground truth mask, respectively, and \textit{k} is the number of labels. 

Performance of various state-of-the-art neural network segmentation models such as U-net, DeepLabV3, DeepLabV3+, PSP, and FPN was tested on the dataset using the \textit{mIoU} metric, and the masks generated from each of the models were compared against the ground truth masks. The segmentation model with high \textit{mIoU} and low loss values was selected for subsequent thermal analysis.

\subsection{Analysis}
The thermal images and the generated masks using the segmentation model were used to identify hot and cool spots in urban areas. The radiometric data collected using the thermal camera was converted to temperature values using the Plank's constants mentioned in Table \ref{tab:camera constants} using the following expression:

\begin{equation}
\label{eq: Ir_response}
T_{obj} = \frac{B}{ln(\frac{R_1}{R_2(U_{tot}+O)}+f)}
\end{equation}
where $U_{tot}$ is the signal response to the long wave infrared radiation incident on the camera detector, \textit{B}, $R_{1}$, $R_{2}$, \textit{O} and \textit{F} are the camera calibration constants. 

FlirExtractor (\cite{klink2019aloisklink}), a Python package, was used for the conversion of radiometric data to temperature values using equation \ref{eq: Ir_response}. Subsequently, the temperature of the region of interest was extracted using the generated mask. The temperature values were then corrected for emissivity depending on the type of urban feature using the following expression (\cite{nichol2009emissivity}):

\begin{equation}
\label{eq: emissivty}
T_{corr} = \frac{T_{obj}}{\sqrt[4]{\epsilon}}
\end{equation}
where, $T_{corr}$ is the corrected temperature value for emissivity $\epsilon$.

Following the temperature correction for emissivity, statistical measures such as mean, median, and standard deviations in the temperature values for each urban feature were estimated at various instances of time. Further, a study was conducted on identifying hot and cool regions in the urban feature. In the subsequent section, the results from the analysis are presented in detail.

\begin{figure}
  \centering  
  \includegraphics[width=1.\linewidth]{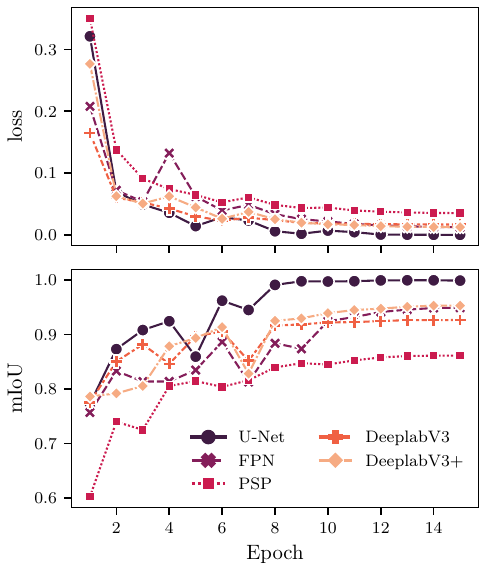}
  \vspace{-5mm}
  \caption{Comparison of loss estimated on the training set (top) and \textit{mIoU} score estimated on the validation dataset (bottom) for five deep learning segmentation models.}
  \label{fig:loss and mIoU}
\end{figure}

\begin{figure}
  \centering  
  \includegraphics[width=1.\linewidth]{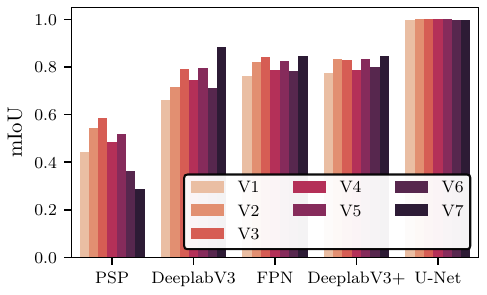}
  \caption{Comparison of \textit{mIoU} score estimated on the test dataset for five deep learning segmentation models for seven views (V1 - V7) in the IRIS dataset. Overall, \textit{U-Net} achieved the highest mean \textit{mIoU} score across all seven views.}
  \label{fig:test-mIoU}
\end{figure}

\begin{table*}[ht!]
  \centering
  \caption{Segmentation model performance on the train and validation set using \textit{`resnet34'} as the encoder.}. 
  \label{tab:nn_performance}
  \begin{tabular}{lcccccc}
    \toprule
     \multirow{2}{*}{Model} &\multicolumn{3}{c}{Train} &\multicolumn{3}{c}{Validation}\\  
     &{Loss} &{Accurary} &{mIoU} &{Loss} &{Accurary} &{mIoU}\\
     \midrule
    \texttt{U-net} & 0.0006 & 0.9999 & 0.9987 & 0.0030 & 0.9996 & 0.9989 \\
    \texttt{DeeplabV3+} & 0.0126 & 0.9951 & 0.9385 & 0.0100 & 0.9963 &  0.9525  \\    
    \texttt{FPN} & 0.0126 & 0.9951 & 0.9375 & 0.0176 & 0.9957 & 0.9488 \\
    \texttt{DeeplabV3} &0.0168 & 0.9933 & 0.9197 & 0.0157 & 0.9939 & 0.9263  \\    
    \texttt{PSP} & 0.0353 & 0.9863 &  0.8562 &  0.0353 & 0.9869 & 0.8610 \\
    \bottomrule
  \end{tabular}
\end{table*}

\section{Results} 
\label{sec:results}
Figure \ref{fig:mask_and_overlay} shows some of the thermal images from the IRIS dataset and the corresponding masks and overlay generated using the LabelMe annotation tool. It can be observed from the masks that the urban features buildings, vegetation, roads, sky, and offshore structures, which are displayed in green, cyan, blue, red, and pink colors, respectively. The final dataset consists of 8,953 images and their corresponding masks. This dataset is split into a test, validation, and train set, which is used to train and test various neural network segmentation models, respectively. The model testing and performance results are discussed in the following section.

\subsection{Segmentation model performance}
Figure \ref{fig:loss and mIoU} shows the loss and the \textit{mIoU} estimated after each epoch on the training and validation set respectively for various state-of-the-art deep learning segmentation models with `resnet34'(\cite{he2016deep}) as the encoder CNN backbone. A constant learning rate of 0.001 was used for training all the models for 15 Epochs. It can be observed from the figure that the loss decreases with the number of epochs and reaches a constant value at the end of ten epochs for the segmentation models. Table~\ref{tab:nn_performance} summarises the performance of five different neural network models on training, validation, and test sets. All the models demonstrate high accuracy and \textit{mIoU} value and a low loss on the validation data. However, on the validation set, the U-net model has the lowest loss value of 0.0030 and highest \textit{mIoU} and accuracy values of 0.9989 and 0.9996, respectively. The differences in the performance amongst various models are a result of variation in the network architecture. In the U-net model, features from each convolution block are used and concatenated with the respective deconvolution block. This type of architecture seems to perform better on the IRIS dataset compared to other neural network architecture.

From the table, it can be seen that the accuracy value is higher than that of \textit{mIoU} value for all the segmentation models. \textit{mIoU} is estimated as the average value of intersection over the union of the pixels found in the prediction and ground truth mask for all labels. At the same time, accuracy is the measure of pixels in the image that are correctly classified. Hence, when the class representation is small, the accuracy metric can be misleading, and \textit{mIoU} represents the model performance. Table\ref{tab:test_results} compares the \textit{mIoU} of five models on the test dataset.  Besides a high accuracy on the validation set, the U-net model also has the highest \textit{mIoU} score of 0.9990 on the test set.  

\begin{figure*}[h]
  \centering
  \includegraphics[width=.8\linewidth]{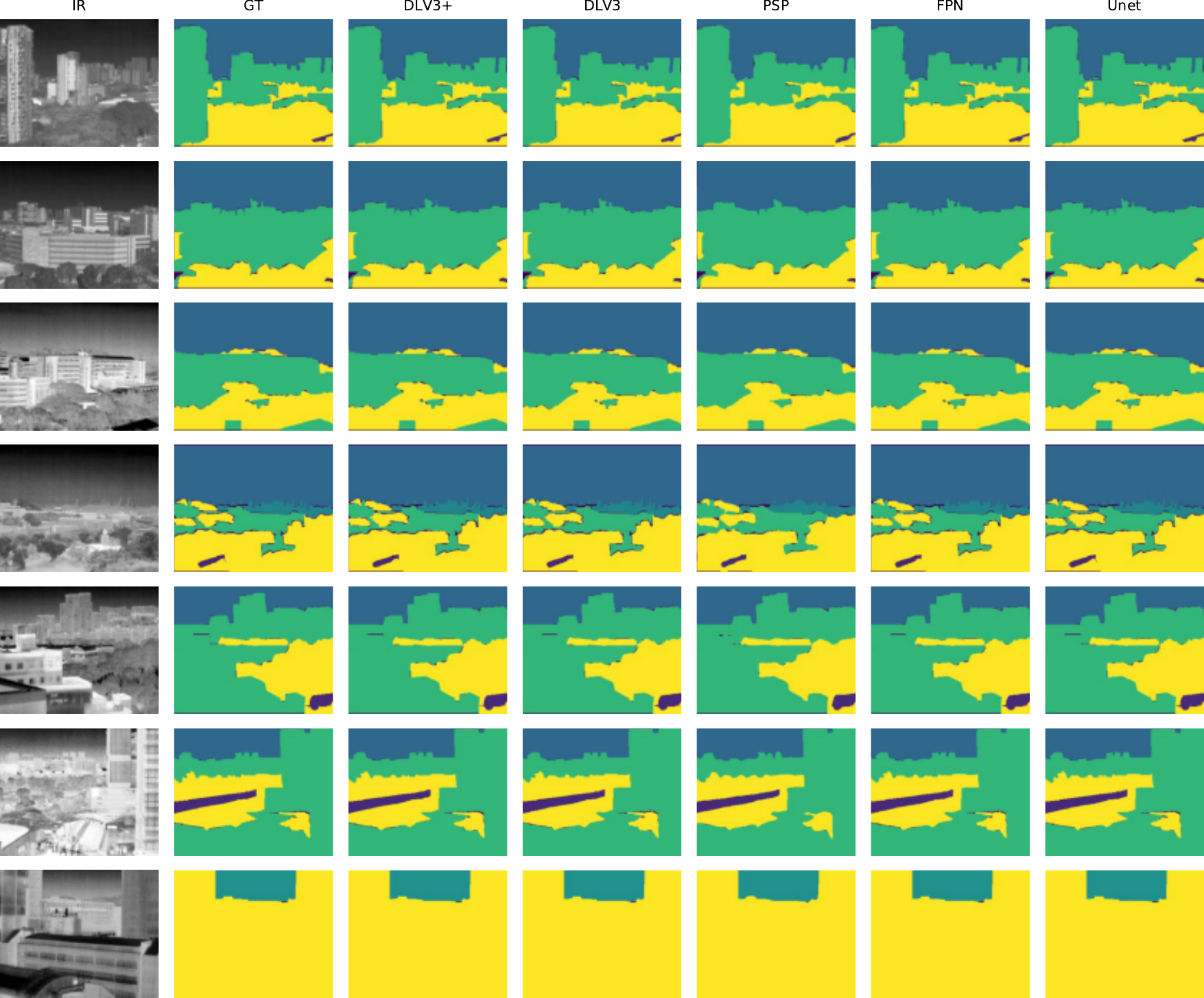}
  \caption{Plot showing the thermal image, ground truth mask, and the masks predicted using various deep learning segmentation models (IR=InfraRed, GT=Ground truth, DLV3=DeeplabV3, DLV3+=DeeplabV3+).}
  \label{fig:Mask_models}
\end{figure*}

Figure \ref{fig:Mask_models}, shows the thermal image, the ground truth mask, and the masks generated using different neural network segmentation models for some of the thermal images. As observed from the Figure, most of the neural network models yield masks of the thermal image that closely represent the ground truth masks. However, the mask generated using the U-net model is more accurate than the others and can also capture the boundaries between different urban features. Thus, for thermal image segmentation, the U-net model is used to generate masks for thermal analysis.

\begin{table*}[ht!]
  \centering
  \caption{Comparison of \textit{mIoU} score estimated on the test dataset for five deep learning segmentation models for seven views in the IRIS dataset. \textit{U-Net} achieved the highest \textit{mIoU} score of 0.9990.}
  \label{tab:test_results}
\begin{tabular}{lrrrrr}
\toprule
Views &     PSP &  DeeplabV3 &     FPN &  DeeplabV3+ &   U-Net \\
\midrule
View 1   &  0.4437 &     0.6636 &  0.7613 &      0.7749 &  0.9992 \\
View 2   &  0.5459 &     0.7172 &  0.8213 &      0.8320 &  0.9998 \\
View 3   &  0.5855 &     0.7906 &  0.8401 &      0.8309 &  0.9999 \\
View 4   &  0.4865 &     0.7467 &  0.7892 &      0.7881 &  0.9998 \\
View 5   &  0.5192 &     0.7957 &  0.8251 &      0.8354 &  0.9999 \\
View 6   &  0.3633 &     0.7114 &  0.7823 &      0.8011 &  0.9956 \\
View 7   &  0.2871 &     0.8858 &  0.8483 &      0.8469 &  0.9989 \\
\midrule
Mean &  0.4616 &     0.7587 &  0.8097 &      0.8156 &  0.9990 \\
\bottomrule
\end{tabular}
\end{table*}

\subsection{Analysis of segmented images}
\begin{figure*}
  \centering  
  \includegraphics[width=.8\linewidth]{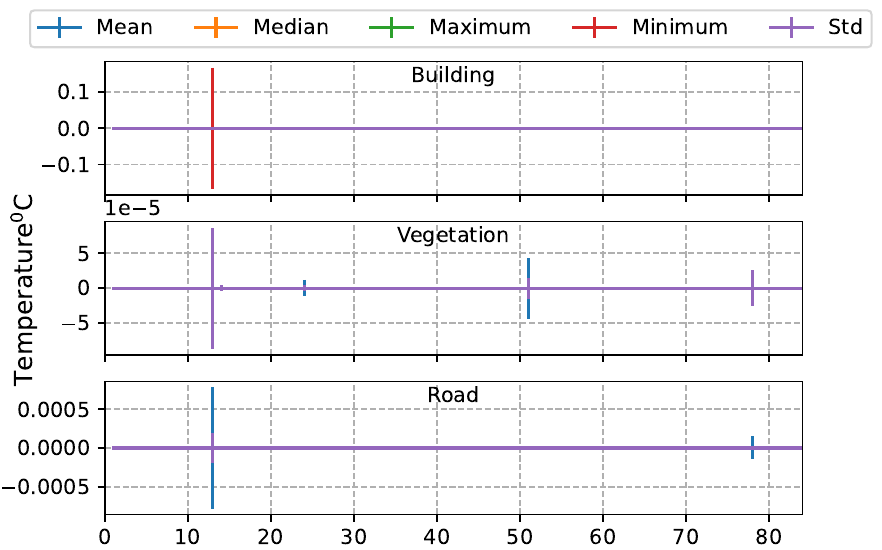}
  \caption{Plot showing the error in the mean, median, minimum, maximum and standard deviation of the temperature extracted using the ground truth mask and the mask generated using deep learning model for various urban features.}
  \label{fig:Temp_accuracy}
\end{figure*}

Evaluation metrics such as \textit{mIoU} scores are usually used to check the performance of the segmentation models. It indicates how the masks generated using the trained segmentation model closely represent the ground truth mask. However, for thermal images, it is essential to check the accuracy of the temperature extracted from the masked features. Various statistical measures of the temperature extracted using the masks generated by the neural network model are compared against the temperature extracted using the ground truth mask and are shown in Figure \ref{fig:Temp_accuracy}. The error in the mean, median, maximum, minimum, and standard deviation in temperature of the urban feature building, vegetation, and road for over eighty images are shown in the figure to demonstrate the deviation in the estimated statistical measure. It can be observed that the error in the mean and the median values of the temperature of the three urban features is close to zero. At the same time, deviation in the maximum and minimum temperature values is observed for some of the images. The highest difference is observed for vegetation in the range of -5 to 5$^{\circ}$C. This discrepancy could be due to the deviations in the edges predicted using the neural network model. Thus, it is noted that the level of accuracy in thermal analysis for any application will depend on the statistical measure to be used. Nevertheless, this discrepancy is observed only for a few images; hence, the trained neural network model can be used to analyze the thermal images further.

\begin{figure*}
	\centering
	\begin{subfigure}{1\columnwidth}
		\includegraphics[width=.425\linewidth]{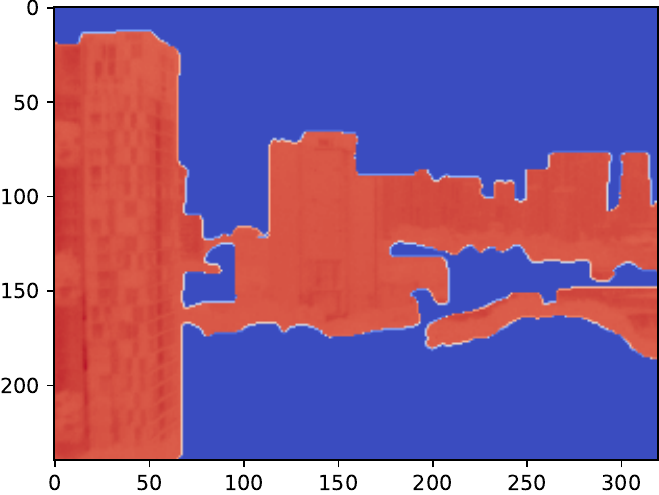}
        \includegraphics[width=.5\linewidth]{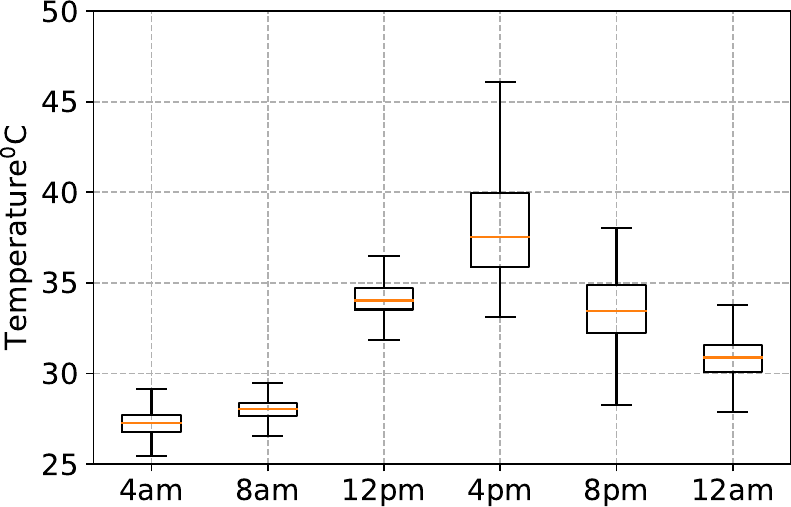}
		\caption{View 1 - building}
		\label{fig:sub-first-a}
	\end{subfigure}
	\begin{subfigure}{1\columnwidth}
		\includegraphics[width=.425\linewidth]{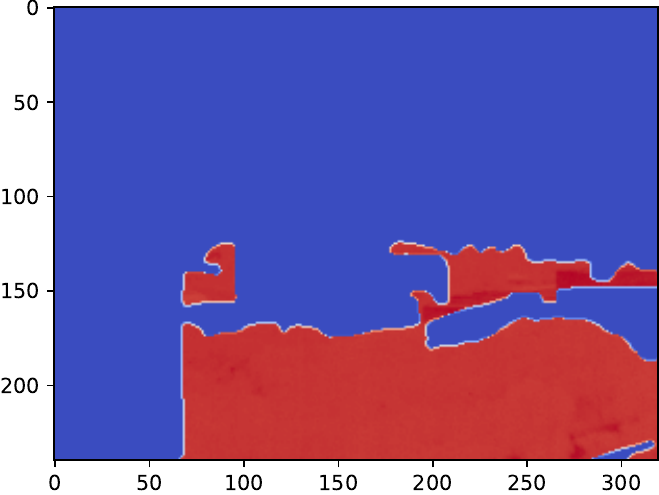}
        \includegraphics[width=.5\linewidth]{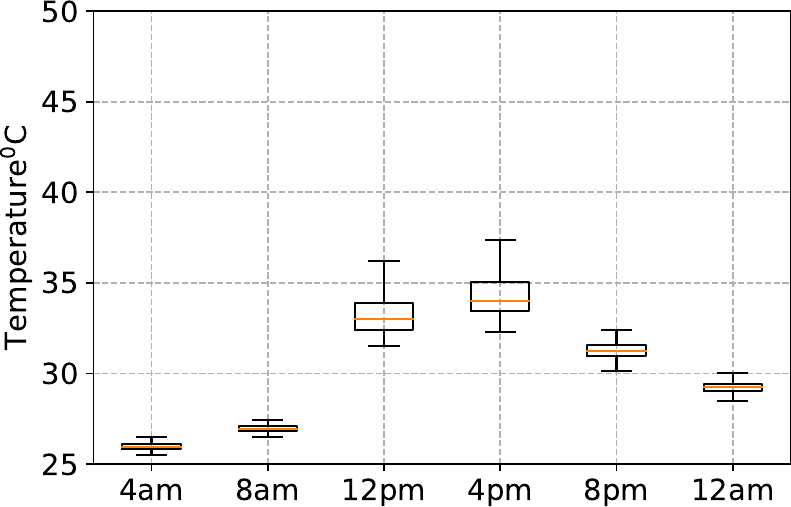}
		\caption{View 1 - vegetation}
		\label{fig:sub-first-b}
	\end{subfigure}
    \hfill
	\begin{subfigure}{1\columnwidth}
		\includegraphics[width=.425\linewidth]{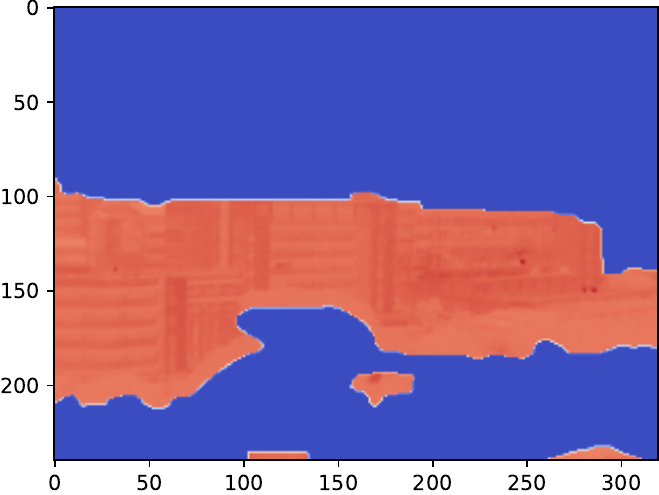}
        \includegraphics[width=.5\linewidth]{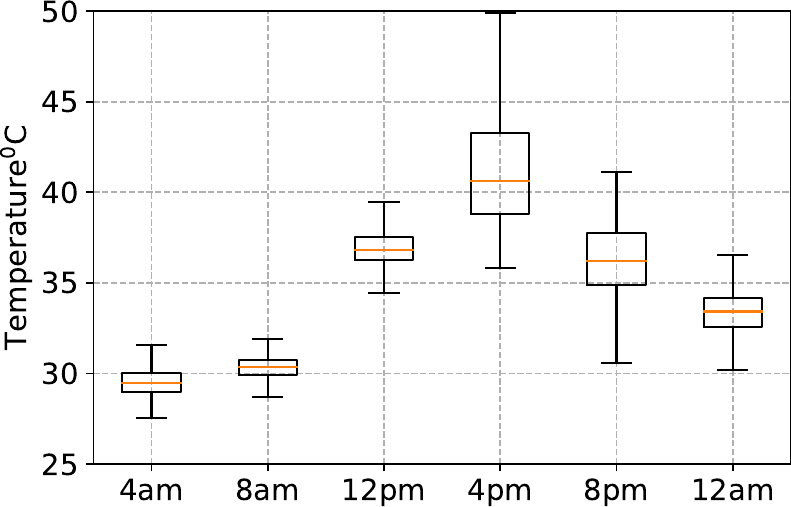}
		\caption{View 3 - building}
		\label{fig:sub-first-c}
	\end{subfigure}
    	\begin{subfigure}{1\columnwidth}
		\includegraphics[width=.425\linewidth]{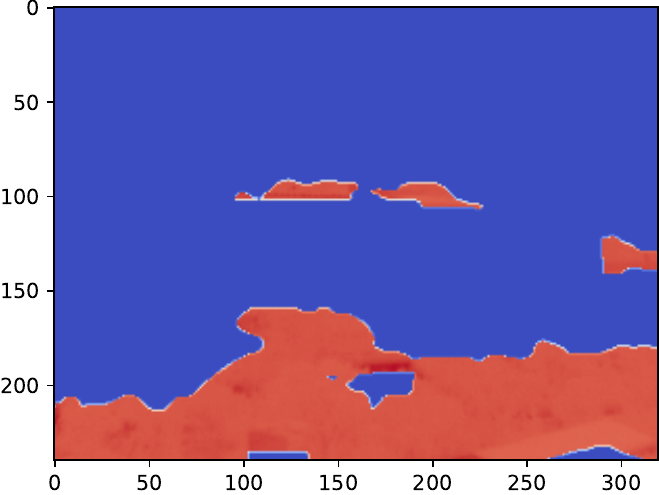}
        \includegraphics[width=.5\linewidth]{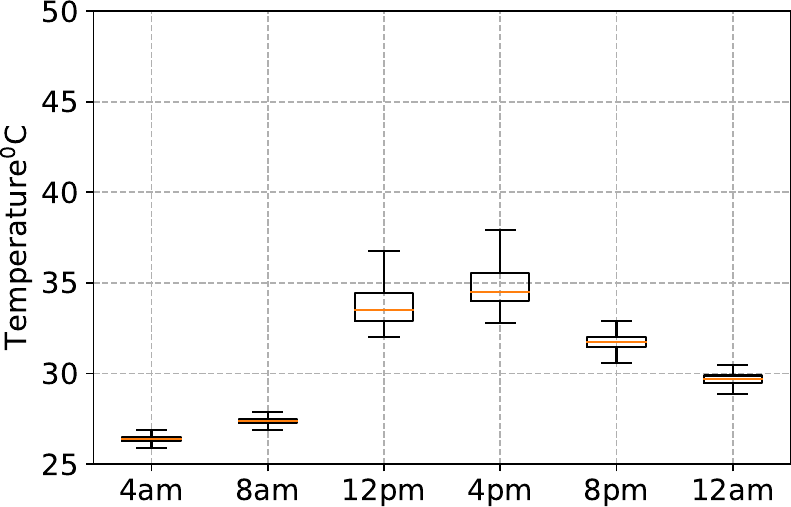}
		\caption{View 3 - vegetation}
		\label{fig:sub-first-d}
	\end{subfigure}	
	\label{fig:subfigures}
\caption{Plot showing the predicted mask of the buildings and vegetation and its corresponding temperature distribution in view 1 (top row) and view 3 (bottom row) (shown in Figure \ref{fig:observatory and location kv}). The box plot shows the dispersion in the temperature of the urban feature at various instances of time in a day.}
\label{fig:Temp_dist}
\end{figure*}

Figure \ref{fig:Temp_dist} shows the predicted mask of the building and vegetation in View 1 and View 3 (shown in Figure~\ref{fig:observatory and location kv}) and its corresponding temperature distribution. The temperature is extracted from the radiometric data at different instances of the day using the masks generated using the U-net segmentation model and corrected for emissivity using Equation \ref{eq: emissivty}. As expected, the temperature of buildings and vegetation for both views is higher during the day than at night due to exposure to solar radiation. Also, the median temperature of buildings and vegetation in both views are comparable. Further, for both views, the buildings' median temperature is higher than that of the surrounding vegetation. Such temporal analysis of the changes in the temperature of the built environment and the surrounding vegetation is essential to devise suitable measures to mitigate the UHI effect in urban areas (\cite{martin2022iris}).

\begin{figure*}
\centering
\begin{subfigure}{1\columnwidth}
  \centering
  \includegraphics[width=1\columnwidth]{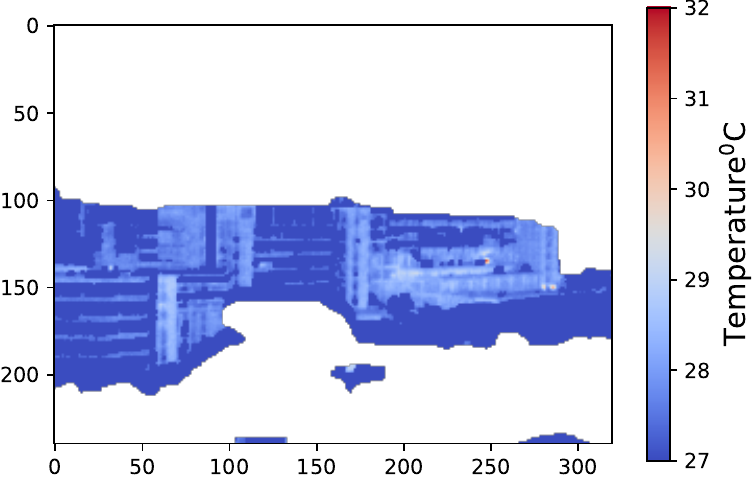} 
  \caption{4am}
  \label{fig:sub-first-2}
\end{subfigure}
\begin{subfigure}{1\columnwidth}
  \centering
  \includegraphics[width=1\columnwidth]{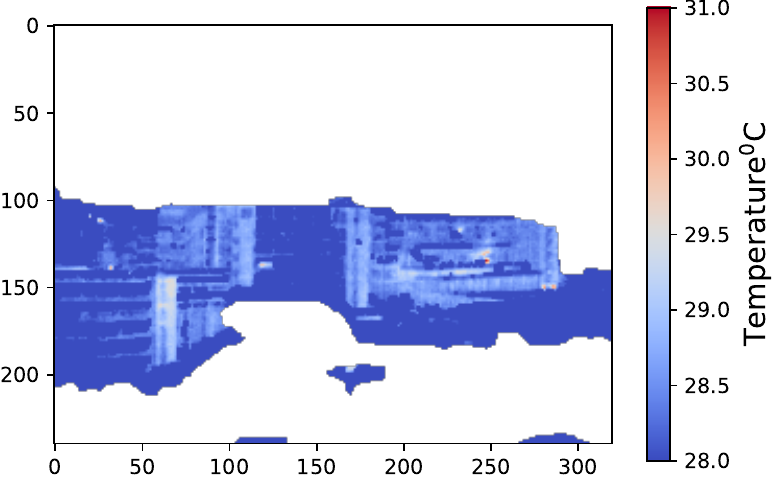} 
  \caption{8am}
  \label{fig:sub-second-2}
\end{subfigure}
\newline
\begin{subfigure}{1\columnwidth}
  \centering
  \includegraphics[width=1\columnwidth]{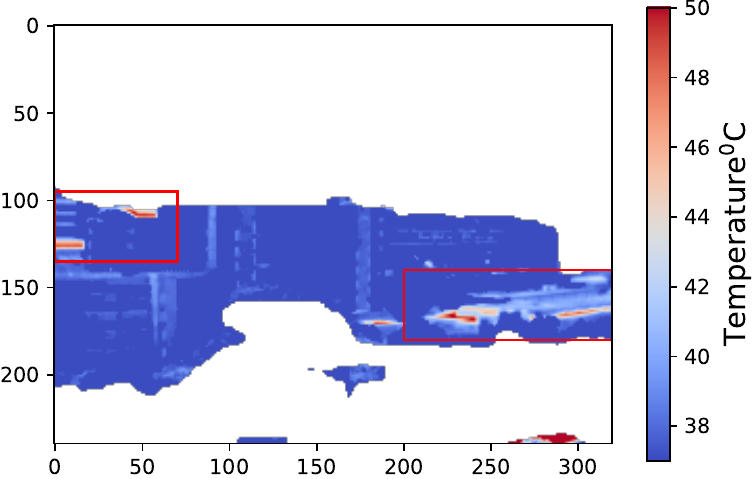}
  \caption{12pm}
  \label{fig:sub-third-2}
\end{subfigure}
\begin{subfigure}{1\columnwidth}
  \centering
  \includegraphics[width=1\columnwidth]{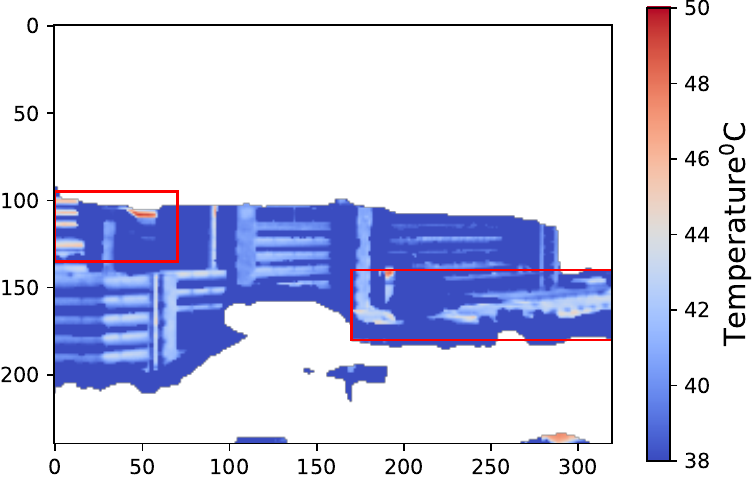}
  \caption{4pm}
  \label{fig:sub-fourth-2}
\end{subfigure}
\newline
\begin{subfigure}{1\columnwidth}
  \centering
  \includegraphics[width=1\columnwidth]{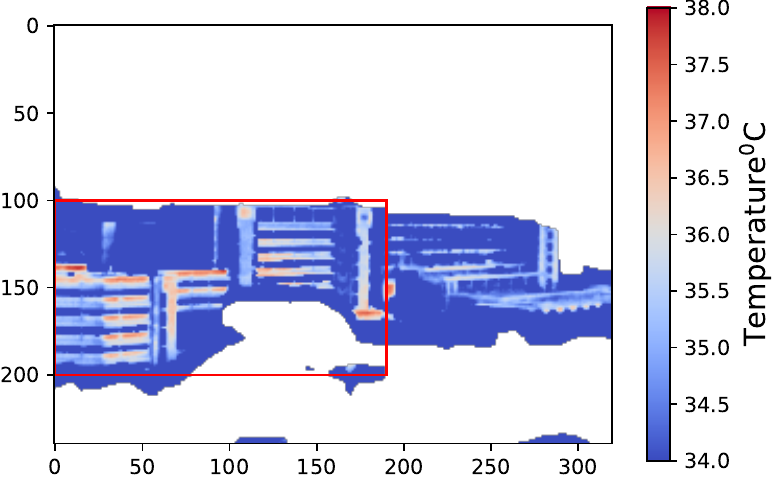}
  \caption{8pm}
  \label{fig:sub-fifth-2}
\end{subfigure}
\begin{subfigure}{1\columnwidth}
  \centering
  \includegraphics[width=1\columnwidth]{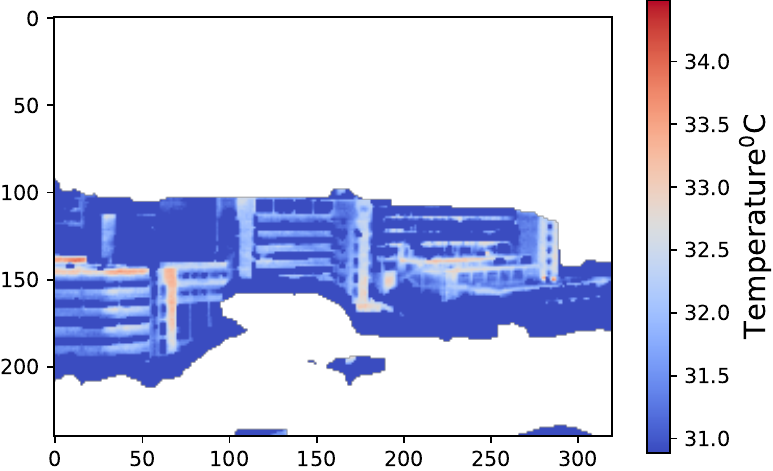} 
  \caption{12am}
  \label{fig:sub-sixth-2}
\end{subfigure}
\caption{Plot showing the hot and cool spots during different times of the day for the buildings in View 3. The hot spots in the region are marked in a red box.}
\label{fig:Temp_hot_cool_spot}
\end{figure*}

It is to be noted that the dispersion in the temperature of the buildings for both views is higher compared to that of vegetation. Also, the dispersion in temperature is higher during the daytime compared to nighttime. To further understand the variation in the temperature spatially, regions with a temperature higher than the mean temperature at various instances of time within one day are shown in Figure \ref{fig:Temp_hot_cool_spot} for the buildings in View 3 (Figure \ref{fig:Temp_dist}(c)). It can be seen that during the early morning, the temperature of various regions of the building is almost uniform. However, at noon, the temperature of exposed roofs and solar panels on the roofs (as shown in Figure \ref{fig:Temp_hot_cool_spot}(c) using a red box) are observed to have higher temperatures compared to the rest of the building space. These exposed roofs can be a potential location for adopting mitigation measures such as green roofs and solar reflective/cool paints (\cite{yang2018green, elnabawi2023numerical}).

Similarly, towards the late evening, the reinforced concrete walls and facade of the buildings (also shown using the red box in Figure \ref{fig:Temp_hot_cool_spot}(e)) have a higher temperature profile compared to the rest of the regions of the building such as the windows and the corridors. The changes in the temperature profile directly affect the heat transfer through the buildings and the surrounding environment and, thereby, the heating and cooling demand (\cite{santamouris2014energy, yang2012integrated}). Besides this, it is essential to identify the hot and the cool spots to evaluate the contributors and mitigators to the day-time and night-time urban heat island effect (\cite{chen2020roles, zhang2017optimizing}). In conclusion, pixel-level identification of hot and cool spots at the neighborhood scale in the urban environment is essential for devising suitable measures to reduce the UHI effect, improve building energy performance, and maximize the thermal comfort of urban dwellers. However, it is noted that surface temperature is not sufficient to quantify the UHI effect as it depends on how much heat penetrates and radiates to the outdoor environment. Thus, an analysis combining the pixel level temperature measure with the indoor and outdoor environmental sensor data will provide a holistic view of the UHI effect in urban areas. 

\begin{figure*}
\centering
\begin{subfigure}{1\columnwidth}
  \centering
  \includegraphics[width=1\columnwidth]{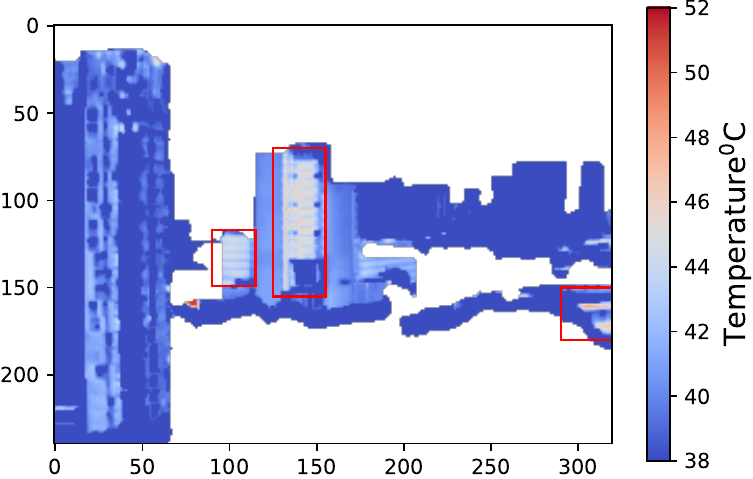} 
  \caption{November'21, 3pm}
  \label{fig:sub-first}
\end{subfigure}
\hfill
\begin{subfigure}{1\columnwidth}
  \centering
  \includegraphics[width=1\columnwidth]{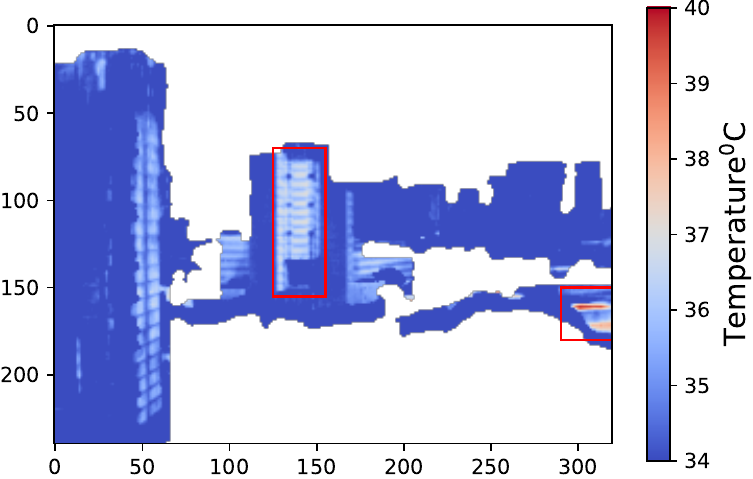} 
  \caption{December'21, 3pm}
  \label{fig:sub-second}
\end{subfigure}
\newline
\begin{subfigure}{1\columnwidth}
  \centering
  \includegraphics[width=1\columnwidth]{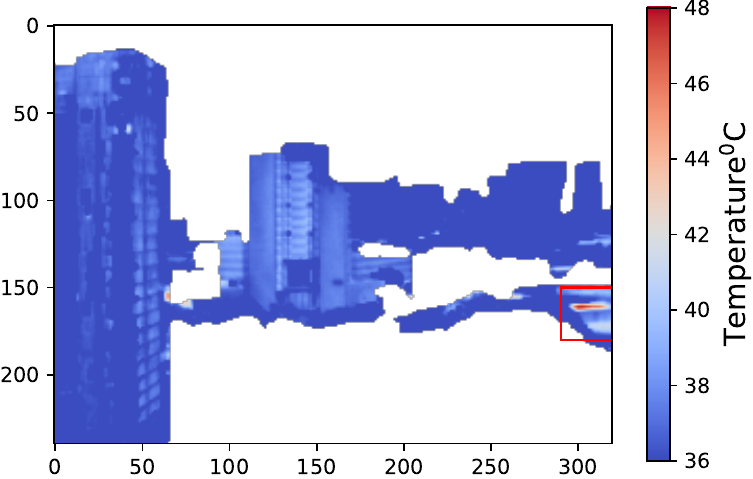}
  \caption{January'22, 3pm}
  \label{fig:sub-third}
\end{subfigure}
\hfill
\begin{subfigure}{1\columnwidth}
  \centering
  \includegraphics[width=1\columnwidth]{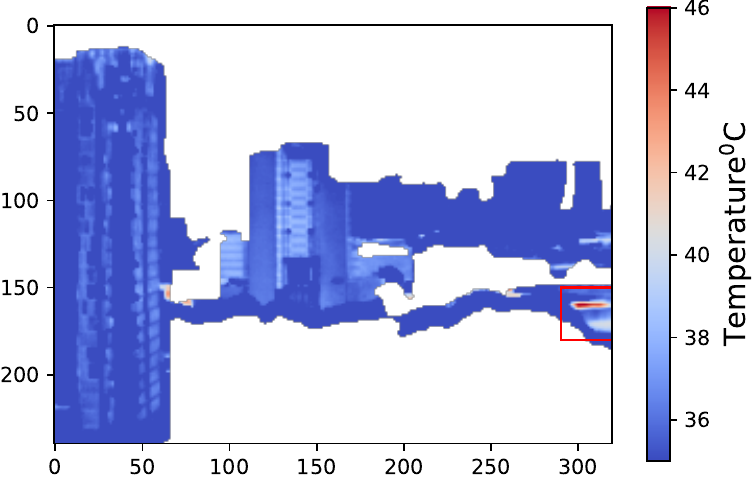}
  \caption{February'22, 3pm}
  \label{fig:sub-first}
\end{subfigure}
\caption{Plot showing the hot and cool regions during four different months at 3 p.m. for the buildings in View 1. The hot spots in the region are marked in red boxes.}
\label{fig:longitudinal_3pm}
\end{figure*}

One of the interesting features of the IRIS dataset is its richness in the temporal domain. Figure \ref{fig:longitudinal_3pm} shows the hot and cool regions of the buildings in view 1 for four different months, captured at 3 p.m. on a clear day. The buildings in View 1 are exposed to intense solar radiation between 10 a.m. to 4 p.m. It is observed that the glass facade of the building and the exposed terrace are identified as hot regions in November and December. In comparison, the exposed terrace is identified as a hot region in January and February. The changes in the sun's path and, thereby, the exposure to solar radiation can result in differences in the identified hot spot across various months. Thus, a long-term study over a few years can be used to gain insights into the impact of various building features on the surrounding environment and monitor the microclimate changes due to climate change.

\section{Discussion}
\label{sec:discussion}
The importance of thermal imaging at the neighborhood scale for urban planning has been demonstrated through various studies such as on heat source detection (\cite{dobler2021urban}), U-value estimation, (\cite{tejedor2019u}), quantifying UHI effect (\cite{martin2022infrared}), building energy audit (\cite{lucchi2018applications}) and operational pattern of the HVAC system (\cite{ramani2023longitudinal}). However, there are very few thermal image datasets at the neighborhood scale; more such thermal datasets are anticipated to be generated, considering the need to monitor the urban environment. Hence, automation of thermal image analysis is important, as this can save time and effort for urban planners. Segmentation models trained on existing neighborhood scale thermal image datasets and the framework described here can offer a starting point for such an analysis. 

\begin{figure}
     \centering
     \begin{subfigure}[b]{.9\columnwidth}
         \centering
         \includegraphics[width=1.\columnwidth]{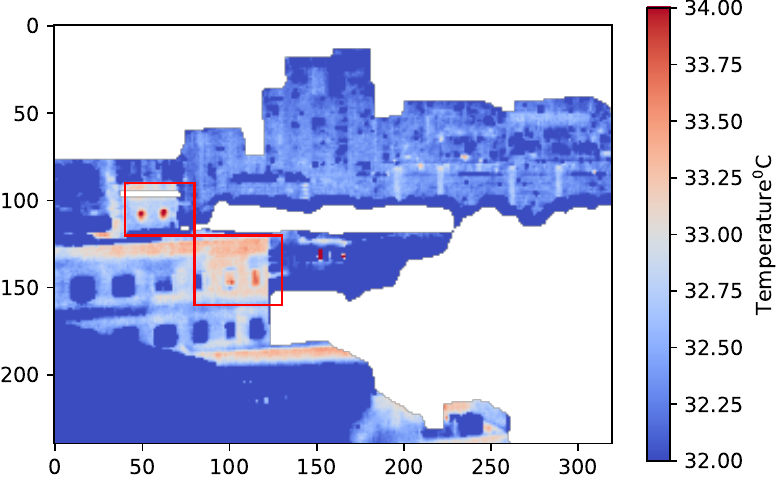}
         \caption{View 5}
         \label{fig:view1}
     \end{subfigure}
     \hfill
     \begin{subfigure}[b]{0.9\columnwidth}
         \centering
         \includegraphics[width=1.\columnwidth]{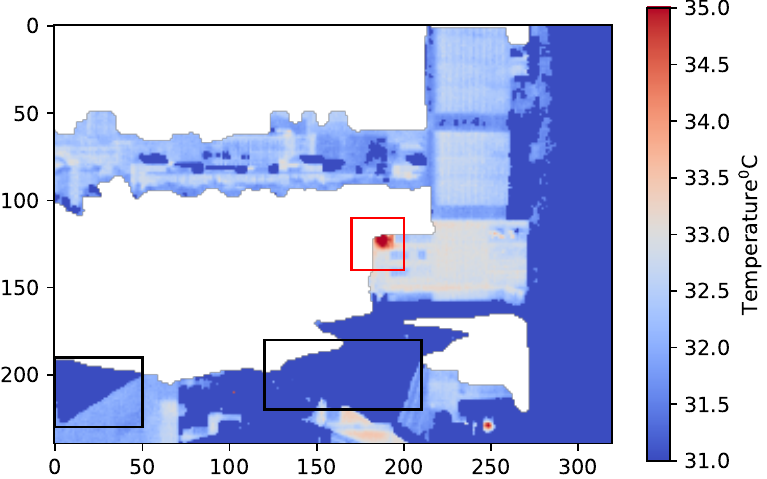}
         \caption{View 6}
         \label{fig:view2}
     \end{subfigure}
     \hfill
     \begin{subfigure}[b]{0.9\columnwidth}
         \centering
         \includegraphics[width=1.\columnwidth]{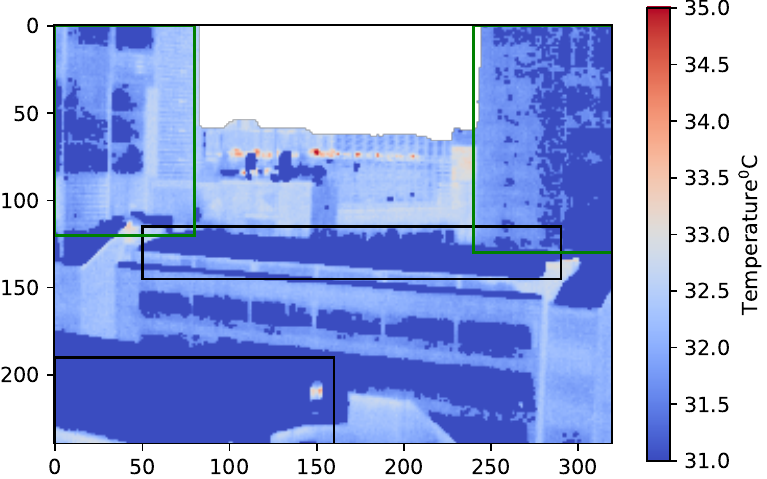}
         \caption{View 7}
         \label{fig:view3}
     \end{subfigure}
        \caption{Plot showing the masked thermal image of the buildings (location 2 shown in Figure \ref{fig:location s16} at 12 am. The red boxes in view 5 and view 6 are the condenser units of the HVAC system. The black boxes in Views 6 and 7 are the solar panels, which are observed to be cooler than the surrounding built environment. The green boxes in View 7 are tall buildings next to each other, resulting in a street canyon.}
        \label{fig:Temp_hot_cool_spot_s16}
\end{figure}

Besides the need for automation, studying the diurnal and seasonal variation of the surface temperature of the buildings and the vegetation can help urban planners locate the hot and cool spots. This will help in the adoption of suitable mitigation measures to reduce UHI and improve building energy efficiency at a local scale and can be decided based on the location of the hot or cool spot. The complexity of the urban environment and the need to identify hot and cool spots can be explained through Figure \ref{fig:Temp_hot_cool_spot_s16}. The red boxes shown in view 5 and view 6 (Figure \ref{fig:Temp_hot_cool_spot_s16}(a) and (b), respectively) are the condenser units of the HVAC system. The heat released from the operation of the condenser units can increase the temperature of the surrounding environment, resulting in increased cooling demand (\cite{yuan2022impact}). Besides the condenser units, the solar panels on the roofs of buildings in view 6 and view 7 (Figure \ref{fig:Temp_hot_cool_spot_s16}(b) and (c) respectively) appear cooler than the rest of the built environment at night time, while these are a hot spot during the day time as shown in Figure \ref{fig:Temp_hot_cool_spot}. The solar roofs can provide shading against direct solar radiation (\cite{wang2020method}); however, they may have high sensible heat flux during the day (\cite{scherba2011modeling}). Another important aspect is the orientation and proximity of the buildings to each other. In view 7 (Figure \ref{fig:Temp_hot_cool_spot_s16}(c)), the two buildings marked using green box are close to each other, which may result in heat accumulation due to the street canyon effect (\cite{karimimoshaver2021effect}), which can impact the space cooling demand (\cite{allegrini2012influence}). The study presented here indicates the complexities involved in the micro-climate analysis of the urban environment. In the future, it aims to conduct an extensive analysis of the thermal images along with the outdoor and indoor environmental monitoring sensor data for an accurate understanding of the diurnal and seasonal changes in the urban environment.

\section{Conclusions}
\label{sec:conclusions}
This paper presents a qualitative analysis of the semantically segmented thermal images to identify hot and cool spots in urban areas. Longitudinal and spatially rich thermal images (\cite{lin2023district}) were collected on the National University of Singapore educational campus for a few months. The thermal images were segmented based on urban features such as buildings, roads, sky, and vegetation. Various state-of-the-art deep learning segmentation models were trained on the IRIS dataset. Based on the model performance, the U-net model has the highest \textit{mIoU} and lowest loss values of 0.9989 and 0.0030, respectively, on the validation data and the highest \textit{mIoU} score of 0.9990 on the test set. It is noted that even though the models demonstrate a high accuracy, there exist a few limitations. For instance, its performance on new thermal images has not been tested. However, these limitations can be addressed by training the models on more such urban-scale thermal image datasets. 

Analysis of the temperature extracted from the segmented images shows that the various statistical measures match well with the temperature extracted from the ground truth masks. Finally, the temperature distribution in the urban feature is analyzed to identify hot and cool spots. A pixel-wise identification of changes in temperature with time like this is essential for evaluating day and night-time UHI in urban areas. This forms one of the few studies aimed at segmenting thermal images using state-of-the-art deep learning models for neighbourhood-scale study of the built environment. In the future, the models presented here can be extended to estimate the pixel-level variation of the heat flux and the contribution of the built environment, vegetation, and other human activities to the UHI effect. Further, a finer segmentation of the various features can be explored to study the temporal and spatial variation of the temperature and its impact on the surrounding environment.

\section*{Acknowledgement}
This research has been supported by the Republic of Singapore’s National Research Foundation through a grant to the Berkeley Education Alliance for Research in Singapore (BEARS) for the Singapore-Berkeley Building Efficiency and Sustainability in the Tropics (SinBerBEST) Program. BEARS has been established by the University of California, Berkeley, as a center for intellectual excellence in research and education in Singapore.

\bibliographystyle{elsarticle-num-names}

\bibliography{cas-refs}

\end{document}